\pdfoutput=1
\documentclass[11pt]{article}

\usepackage{emnlp2021}
\usepackage{times}
\usepackage{latexsym}
\usepackage{amsmath} 
\usepackage{amssymb} 
\usepackage{graphicx} 
\usepackage{multirow}
\usepackage[T1]{fontenc}

\usepackage[utf8]{inputenc}

\usepackage{microtype}
\usepackage{booktabs}
\title{Description-based Label Attention Classifier for Explainable ICD-9 Classification}

\author{Malte Feucht$^{1}$, Zhiliang Wu$^{2,3}$, Sophia Althammer$^{4}$, Volker Tresp$^{2,3}$\\
  $^{1}$Department of Informatics, TU Munich
  $^{2}$Institute of Informatics, LMU Munich \\
  $^{3}$Technology, Siemens AG
  $^{4}$Faculty of Computer Science, TU Wien \\
  \texttt{malte.feucht@tum.de},
  \texttt{\{zhiliang.wu,volker.tresp\}@siemens.com},\\
  \texttt{sophia.althammer@tuwien.ac.at}\\}

\begin{document}
\maketitle
\begin{abstract}
ICD-9 coding is a relevant clinical billing task, where unstructured texts with information about a patient's diagnosis and treatments are annotated with multiple ICD-9 codes. Automated ICD-9 coding is an active research field, where CNN- and RNN-based model architectures represent the state-of-the-art approaches. In this work, we propose a description-based label attention classifier to improve the model explainability when dealing with noisy texts like clinical notes. 
We evaluate our proposed method with different transformer-based encoders on the MIMIC-III-50 dataset. Our method achieves strong results together with augmented explainablilty.
\end{abstract}

\section{Introduction}
Physicians are obliged to thoroughly document every patient encounter. Structured and semi-structured reports become more common, which contain comprehensive information about performed treatments, procedures and diagnoses. They are typically annotated with multiple billing codes, the international classification of diseases codes (ICD-9 in the US, ICD-10 in Europe). Annotating the noisy discharge summaries with ICD-9 codes is not only manual and labor-intensive, but also error-prone, which has attracted much attention both from clinical and technical perspectives. To facilitate the clinical workflow, we propose a new approach with state-of-the-art annotation performance while providing an explanation for the proposed annotation. 

\citet{de1998hierarchical} introduced automated ICD-9 coding as a text-based multi-label classification problem. Deep learning-based approaches, which exploit convolutional (CNNs) and recurrent neural networks (RNNs) with attention mechanisms~\citep{Shi, mullenbach-etal-2018-explainable,Vu} define the current  state-of-the-art. Meanwhile, large-scale pre-trained language models based on transformer~\citep{NIPS2017_3f5ee243} architectures have demonstrated considerable performance improvements for text-based tasks, e.g., BERT~\citep{devlin-etal-2019-bert}. Especially their ability to model long-range dependencies within an input sequence would potentially benefit the task of ICD-9 coding since the information for a certain label prediction can be distributed across the whole text. Unlike other areas of natural language processing (NLP), little research on applying transformer-based architectures on the task of ICD-9 coding has been explored~\citep{pascual-etal-2021-towards,biswas2021transicd,ji2021does}. \citet{sun2020understanding} argue that attention scores are able to capture global, absolute importance of work tokens and can thus provide some degree of explainability for text classification.

In this work, we propose a \textbf{description-based label attention classifier (DLAC)}. We show that it can be applied to different transformer-based encoders and provides explainable predictions on noisy texts. DLAC learns ICD-9 code embeddings by integrating the descriptions of the ICD-9 codes and applies the embeddings to the respective text representations to obtain label-specific representations for each code classification. We evaluate different model architectures on the MIMIC-III-50 dataset, a benchmark dataset for the task of ICD-9 coding, in order to answer the following research questions (RQs).

\textbf{RQ1:} \textit{Which transformer-based encoder is best suited for ICD-9 coding?}

We evaluate and compare BERT \citep{devlin-etal-2019-bert}, hierarchical BERT \citep{pappagari2019hierarchical} and Longformer \citep{Longformer} as transformer-based encoders and find that the Longformer \citep{Longformer} yields the best results for ICD-9 coding.
Furthermore, we investigate:

\textbf{RQ2:} \textit{How does our proposed description-based label attention classifier perform on ICD-9 coding?}

We compare DLAC with a common logistic regression classifier (LRC) on top of the transformer-based encoder for ICD-9 coding. While adding explainability to the predictions, DLAC outperforms the corresponding LRC by 1-4 \%. 
Since explainable predictions are crucial for noisy texts, like discharge summaries, we investigate:

\textbf{RQ3:} \textit{To which extent can the DLAC provide explainable predictions for ICD-9 codes?}

Since the attention scores in DLAC offer a way to explain the predictions with respect to different text segments, we analyze the top attention scores of DLAC and demonstrate the utility as part of a graphical interface.

\section{Methods}
As a multi-label text classification problem, each discharge summary is represented by a tokenized input sequence of words $\boldsymbol{X}_i:=[\boldsymbol{x}_1, \dots, \boldsymbol{x}_{t_i}]\in\mathbb{R}^{l\times t_i}$, where $l$ denotes the vector dimension and $t_i$ is the length of the $i$-th input sequence. The goal is to predict a binary vector $\boldsymbol{y}_i\in \mathbb{R}^m$, where $m$ represents the set size of ICD-9 codes.
Each element in the predicted vector $\boldsymbol{y}$ is of value $0$ or $1$. In the following, we denote scalars with lowercase letters like $x$, vectors with bold lowercase letters like $\boldsymbol{x}$, and matrices with bold uppercase letters like $\boldsymbol{X}$.

\subsection{Description-based Label Attention}

\begin{figure}[hb]
\centering
\includegraphics[width=0.4\textwidth]{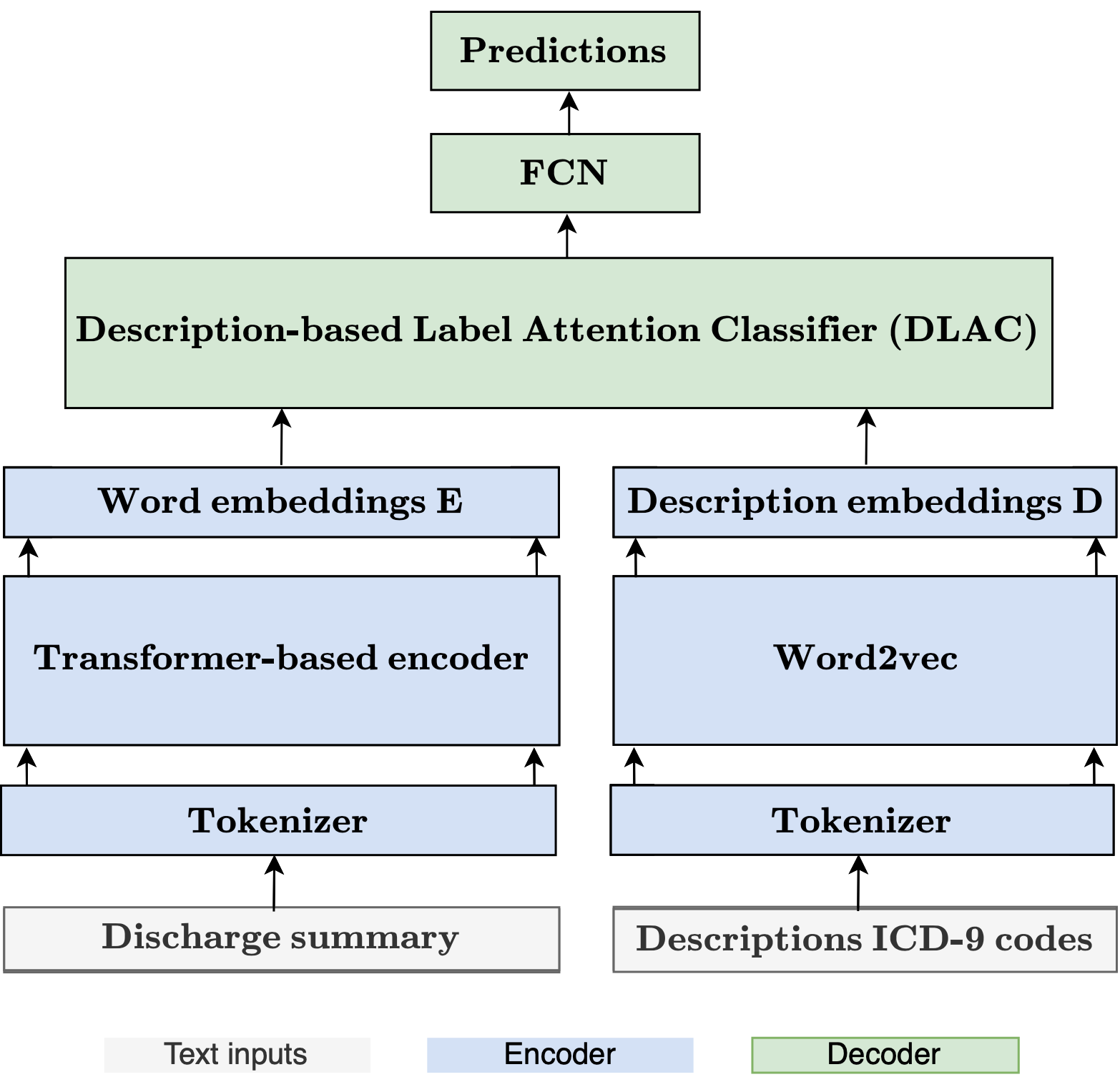}
\caption{Overall model architecture}
\label{fig:overall_model_architecture}
\end{figure}

An overview of the proposed model architecture with DLAC is illustrated in Figure~\ref{fig:overall_model_architecture}. It includes a transformer-based encoder to represent the discharge summaries into a word embedding matrix $\boldsymbol{E}$. Meanwhile, the descriptions of the ICD-9 codes are represented with a description embedding matrix $\boldsymbol{D}$, which is initialized by ICD-9 code descriptions with Word2vec embeddings \citep{mikolov2013distributed}. 

\begin{figure}[t]
\centering
\includegraphics[width=0.48\textwidth]{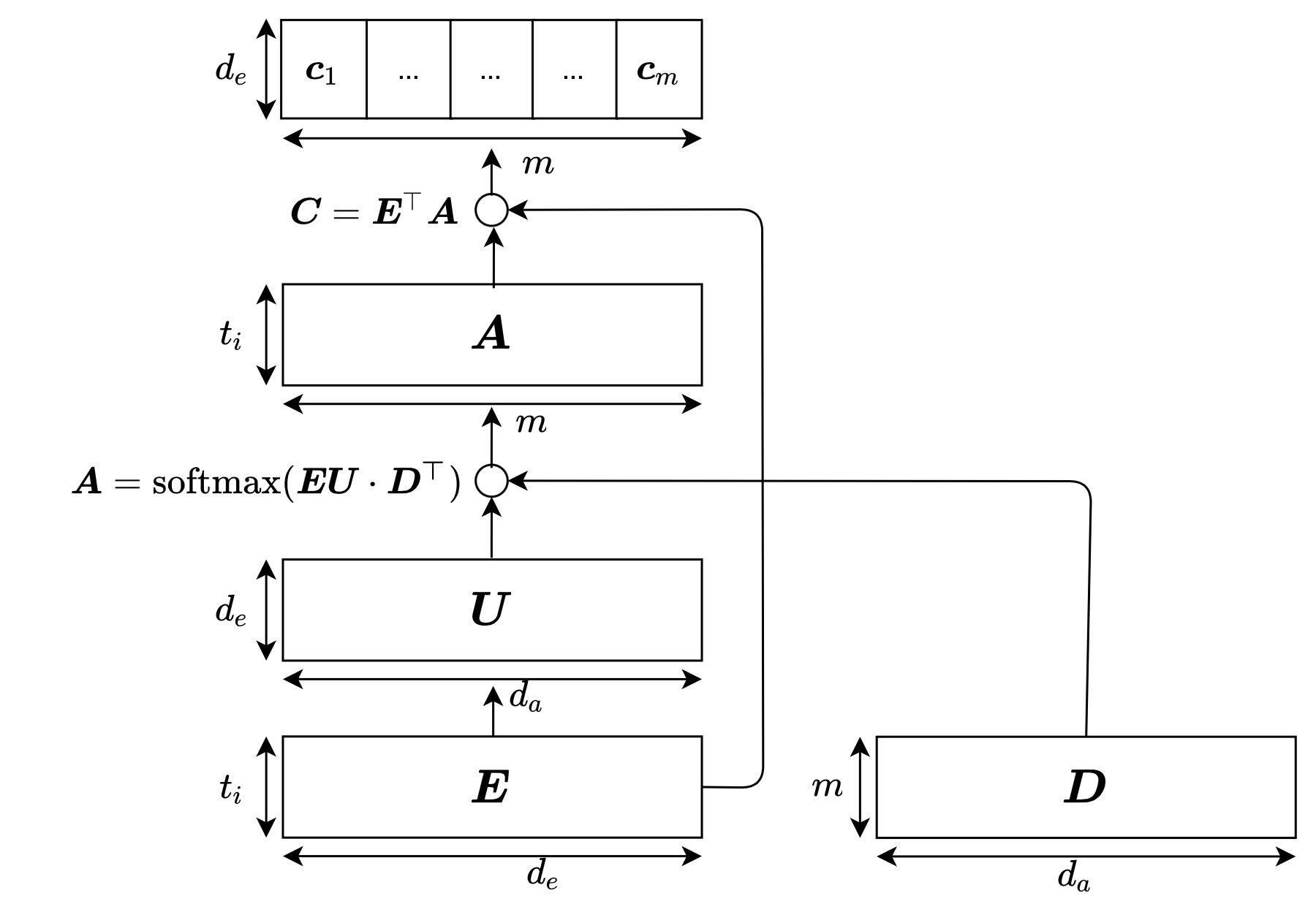}
\caption{Description-based Label Attention Classifier}
\label{fig:lac_detailed}
\end{figure}

More specifically, as shown in Figure~\ref{fig:lac_detailed}, DLAC is fed with two matrices:
The word embedding matrix $\boldsymbol{E}\in\mathbb{R}^{t_{i}\times d_{e}}$ is computed from a transformer-based encoder and the ICD-9 code description embedding matrix $\boldsymbol{D}\in\mathbb{R}^{m\times d_{a}}$. Here, $d_{e}, d_{a}$ denote  the dimension of the word embeddings and the description embeddings, respectively. The description embedding matrix $\boldsymbol{D}$ is set to be trainable. To compute the attention score ${a}_{jk}$ for the $i$-th word on the $j$-th label, we first apply a dimension transformation on the word embedding matrix with  $\boldsymbol{U}\in\mathbb{R}^{d_{e}\times {d_{a}}}$ to match the shape of description matrix $\boldsymbol{D}$. The ICD-9 code description vectors stored in the description matrix $\boldsymbol{D}$ can be seen as queries that include the essential information from the label description of respective ICD-9 codes. Formally, we compute the label attention matrix $\boldsymbol{A}\in\mathbb{R}^{t_{i}\times m}$ as
\begin{equation*}
\boldsymbol{A}=\text{softmax}(\boldsymbol{E}\boldsymbol{U}\cdot \boldsymbol{D^{\top}}).
\end{equation*}

% where each column denotes the attention scores of words for a specific coding label.
After that, we compute contextual embeddings for each label by aggregating information from the word embedding matrix $\boldsymbol{E}$ with attention scores in $\boldsymbol{A}$. More concretely, the contextual embedding matrix $\boldsymbol{C}\in\mathbb{R}^{d_e\times m}$ is computed as
\begin{equation*}
\boldsymbol{C}=\boldsymbol{E}^{\top}\boldsymbol{A}.
\end{equation*}
\subsection{Classification}
Each label specific contextual embedding is fed into a single layer fully-connected network (FCN) for the prediction of the respective ICD-9 code label. A sigmoid activation function is applied to have a probabilistic prediction. The training objective is the binary cross-entropy loss computed from predictions $\boldsymbol{\hat{y}}_i\in\mathbb{R}^{m}$ and ground-truth labels $\boldsymbol{y}_i\in\mathbb{R}^{m}$. 
\subsection{Transformer-based Encoders}
Since DLAC is agnostic to encoders, we conduct experiments with different transformer-based encoders by integrating them into the proposed model architecture.
%as illustrated in figure~\ref{fig:overall_model_architecture}.
The first one is a pre-trained \(\textbf{BERT}_{\textbf{BASE}}\) model~\citep{devlin-etal-2019-bert}, which can only consume input sequences with a length of up to $512$ tokens. Longer discharge summaries are simply truncated. The second architecture is a \(\textbf{hierarchical BERT}_{\textbf{BASE}}\) model \citep{pappagari2019hierarchical}, which aims to overcome the input sequence length limitation. The discharge summary of length $t_{i}$ is split into $k$ overlapping chunks, where $k=\lceil\frac{t_{i}}{512}\rceil$, where $\lceil\cdot\rceil$ is a ceiling operation. The chunks are fed sequentially into the \(\textbf{BERT}_{\textbf{BASE}}\) model to obtain the word embedding matrices for every chunk, which are then averaged using mean-pooling across all chunks. The third architecture is a pre-trained \(\textbf{Longformer}_{\textbf{BASE}}\) model~\citep{Longformer}. A limitation of transformer-based language models such as BERT is their inability to process long input sequences due to the computational cost of the self-attention mechanism, which scales quadratically with the input sequence length. The \(\textbf{Longformer}_{\textbf{BASE}}\) model overcomes this limitation by offering a "sparsified" self-attention mechanism, making it more suitable to process longer input sequences. As a result, the \(\textbf{Longformer}_{\textbf{BASE}}\) model can process input sequences of lengths of up to $4096$ tokens.

\section{Data}
MIMIC-III~\citep{johnson2016mimic} is a large, freely available clinical database. Similar to \citet{mullenbach-etal-2018-explainable}, we create the subset MIMIC-III-50 from the full dataset. It includes discharge summaries containing the $50$ most frequent ICD-9 codes, as otherwise, the label distribution is extremely imbalanced. After pre-processing, MIMIC-III-50 contains $11.368$ samples and $50$ ICD-9 codes, where a summary of different statistics can be found in Table~\ref{tab:descriptive_statistics_MIMIC_3_50}. Following the pre-processing in \citet{mullenbach-etal-2018-explainable}, we first lowercase all tokens, remove punctuations and remove numerical tokens-only.
The MIMIC-III-50 dataset is split into a training, validation and test set with $8.066$, $1.573$ and $1.729$ samples, respectively.
\begin{table}[t]
\centering
\begin{tabular}{llll}
\hline
\textbf{MIMIC-III-50} & \textbf{\# Words} & \textbf{\# ICD-9 Codes}\\
\hline
\textbf{mean} &  1.612  & 5,77\\
\textbf{std}  & 788 & 3,37\\
\textbf{min}  & 105 & 1\\
\textbf{max}  & 7.567 & 24 \\
\textbf{25\%}  & 1.065 & 3\\
\textbf{50\%}  & 1.478 & 5\\
\textbf{75\%}  & 1.992 & 5\\ \hline
\end{tabular}
\caption{Descriptive statistics of MIMIC-III-50 dataset}
\label{tab:descriptive_statistics_MIMIC_3_50}
\end{table}

\section{Experiments}
\begin{table*}
\centering
\begin{tabular}{cccccc}
\hline
\multirow{2}{*}{\textbf{Model}} &
     \multicolumn{2}{c}{\textbf{AUC}} &
     \multicolumn{2}{c}{\textbf{F1}} &{\textbf{P@5}} \\
     & Macro & Micro & Macro & Micro \\
\hline
BERT+LRC  & 0.80 $\pm$ 0.007  & 0.84 $\pm$ 0.006 & 0.33 $\pm$ 0.033 & 0.45 $\pm$ 0.026 & 0.51 $\pm$ 0.011 \\
H-BERT+LRC  & 0.82 $\pm$ 0.006 & 0.86 $\pm$ 0.006 & 0.29 $\pm$ 0.030 & 0.41 $\pm$ 0.032 & 0.51 $\pm$ 0.012\\
Longformer+LRC  & 0.85 $\pm$ 0.003 & 0.89 $\pm$ 0.003  & 0.48 $\pm$ 0.005  & 0.58 $\pm$ 0.003  & 0.59 $\pm$ 0.012 \\
BERT+DLAC & 0.80 $\pm$ 0.006  & 0.84 $\pm$ 0.004 & 0.35 $\pm$ 0.032  & 0.46 $\pm$ 0.026 & 0.51 $\pm$ 0.013\\
H-BERT+DLAC & 0.83 $\pm$ 0.035  & 0.87 $\pm$ 0.004 & 0.32 $\pm$ 0.020 & 0.43 $\pm$ 0.013 & 0.52 $\pm$ 0.010 \\
Longformer+DLAC* & $0.87 \pm 0.008$ & $0.91 \pm 0.006$ & $0.52 \pm 0.020$ & $0.62  \pm 0.024$ &  $0.61 \pm 0.013$\\
\hline\hline
JointLAAT** & $\mathbf{0.93}$ & $\mathbf{0.95} $ & $\mathbf{0.67} $ & $\mathbf{0.72} $ & $\mathbf{0.68}$\\
TransICD & 0.89 & 0.92 & 0.56 & 0.64 &  0.62 \\
\hline
\end{tabular}
\caption{
Test results on the MIMIC-III-50 dataset for all proposed model architectures compared to the state-of-the-art architectures. ** marks the best overall model architecture. * marks the best model architecture of this work.
% * marks the best encoder architecture of our work. ** marks the best overall architecture of our work.
}
\label{tab:test_results}
\end{table*}
We train the proposed description-based label attention classifier (DLAC) using BERT, hierarchical BERT, and Longformer as encoders. As a baseline classifier, we choose a simple logistic regression classifier (LRC) on top of the different encoders. In contrast to DLAC, LRC does not take the description embeddings $\boldsymbol{D}$, into account. We set $d_a=600$ and $d_e=768$. We train all architectures using Adam optimizer with a learning rate of $\alpha=1.41\times 10^{-5}$ and a global batch size of $64$. Furthermore, we use k-fold cross-validation with $k=5$ folds and train every fold for $25$ epochs. For regularization, we use dropout layers with a probability set to $p=0,1$ and early stopping.
We use the widely adopted micro-and macro averaged area under the ROC curve (AUC), F1 and Precision@n as evaluation metrics to ensure comparability with other works.For P@n we choose $n=5$ because this roughly equals the average number of ICD-9 codes one discharge summary is annotated with, which is $5,77$ for the MIMIC-III-50 dataset. The implementation is made available to ensure the reproducibility of the work\footnote{https://git.io/JzOyk}.

\section{Results and Discussions}
Table~\ref{tab:test_results} presents the results for all proposed model architectures. The Longformer+DLAC yields the best results across all metrics for the architectures in this work.
%We compare different aspects and indications of the obtained results for the posed research questions.
\subsection{RQ1: Transformer-based Encoders}
Among the transformer-based encoders that are combined with the simple LRC classifier (BERT+LRC, H-BERT+LRC, Longformer+LRC), the Longformer encoder yields the best results across all metrics. 
This can be attributed to the inability of BERT to process sequences longer than $512$, where over $75\%$ of the discharge summaries are truncated and potentially important information is disregarded.
For the micro- and macro F1 scores, H-BERT+LRC model yields even poorer results than the BERT+LRC model. This indicates that the way of aggregating the chunks $k$ using mean-pooling is suboptimal, and thus the model fails to create rich input feature representations.

\subsection{RQ2: Longformer+DLAC}
The results of the Longformer+DLAC model show that it performs well on the task of ICD-9 coding. In addition, DLAC outperforms the LRC for all encoder architectures on the task of ICD-9 coding across all metrics by $1-4\%$. 
Meanwhile, JointLAAT~\citep{Vu} is one state-of-the-art CNN-based model architecture. Our Longformer+DLAC underperforms it by $\Delta-0.06, \Delta-0.04, \Delta-0.15, \Delta-0.10, \Delta-0.07$ for Macro AUC, Micro AUC, Macro F1, Micro F1 and P@5 respectively. 
In comparison to state-of-the-art transformer-based architectures, like TransICD~\citep{biswas2021transicd}, our model shows a comparable performance with the difference being $\Delta-0.02, \Delta-0.01, \Delta-0.04, \Delta-0.02, \Delta-0.01$ for Macro AUC, Micro AUC, Macro F1, Micro F1 and P@5, respectively. Transformer-based models haven't reached state-of-the-art dominated by more lightweight CNN-based architectures. This can be partially attributed to the fact that the MIMIC-III-50 dataset does not hold enough training samples for training such a large architecture, e.g., the Longformer+DLAC has up to $152$ million trainable parameters. Furthermore, the Longformer+DLAC model could potentially be improved by using a domain-specific, pre-trained Longformer architecture \citep{PubMedBERT} and by developing a regularization mechanism \citep{cao2019learning} that helps to classify rare labels more accurately. On the other hand, as an encoder agnostic classifier, DLAC can be combined with other models to improve the performance while keeping explainable predictions. 

\subsection{RQ3: Explainability}
\begin{figure}[t]
\centering
\includegraphics[width=0.48\textwidth]{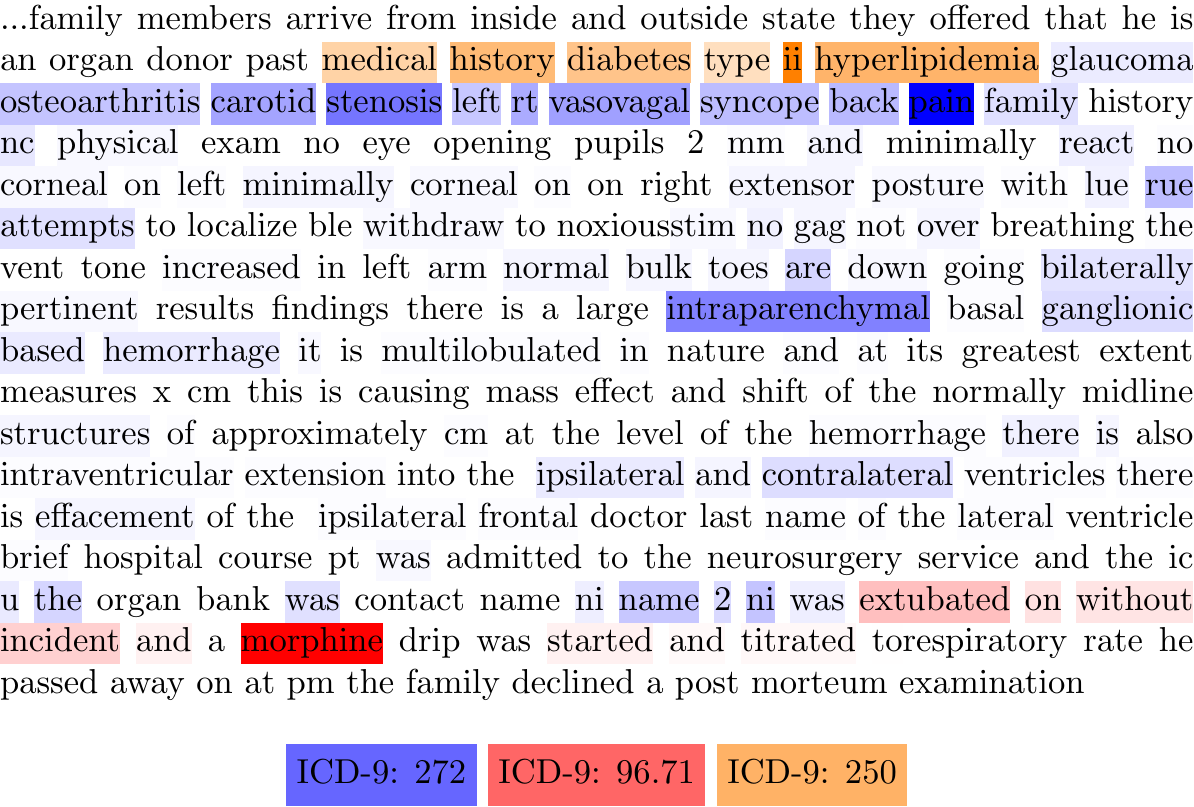}
\caption{The top attention scores for the predicted ICD-9 codes \textbf{272, 96.71, 250} are highlighted with color intensity. Higher color intensity represents larger attention scores and vice versa.}
\label{fig:attention_scores_marker}
\end{figure}

In a setting where a machine learning model would serve as a decision support tool for medical workers, explainability of the obtained model predictions are of utmost importance. In contrast to LRC, DLAC provides explainable predictions using the attention scores. DLAC can retrieve the top attention scores for each ICD-9 code prediction. As an example, the text segments that lead to a certain ICD-9 code prediction are highlighted with respective color intensity in Figure~\ref{fig:attention_scores_marker}.
This can be useful for working with noisy texts in general because it provides some extent of explainability.

\section{Conclusion}
Transformer-based architectures show promising performance on the task of ICD-9 coding. We find that the Longformer encoder is the best suitable encoder architecture for processing long, noisy input sequences such as discharge summaries. We show that our proposed description-based label attention classifier (DLAC) can be applied to various transformer-based encoders and the resulting model outperforms a common decoder architecture like logistic regressions by 1-4\%.
In addition, our proposed DLAC is especially suitable for a practical use case when working with fuzzy, long texts such as the discharge summaries, where explainability for the predicted ICD-9 codes is necessary.

% Entries for the entire Anthology, followed by custom entries
\bibliography{anthology,custom}
\bibliographystyle{acl_natbib}
\end{document}